# Faster Spatially Regularized Correlation Filters for Visual Tracking

Xiaoxiang Hu, Yujiu Yang

Tsinghua University

Shenzhen Key Laboratory of Borad-band Network & Multimedia

Graduate School at Shenzhen, Tsinghua University, Shenzhen, china

## Abstract

*Discriminatively learned correlation filters (DCF) have been widely used in online visual tracking filed due to its simplicity and efficiency. These methods utilize a periodic assumption of the training samples to construct a circulant data matrix, which implicitly increases the training samples and reduces both storage and computational complexity. The periodic assumption also introduces unwanted boundary effects. Recently, Spatially Regularized Correlation Filters (SRDCF) solved this issue by introducing penalization on correlation filter coefficients depending on their spatial location. However, SRDCF's efficiency dramatically decreased due to the breaking of circulant structure.*

*We propose Faster Spatially Regularized Discriminative Correlation Filters (FSRDCF) for tracking. The FSRDCF is constructed from Ridge Regression, the circulant structure of training samples in the spatial domain is fully used, more importantly, we further exploit the circulant structure of regularization function in the Fourier domain, which allows our problem to be solved more directly and efficiently. Experiments are conducted on three benchmark datasets: OTB-2013, OTB-2015 and VOT2016. Our approach achieves equivalent performance to the baseline tracker SRDCF on all three datasets. On OTB-2013 and OTB-2015 datasets, our approach obtains a more than twice faster running speed and a more than third times shorter start-up time than the SRDCF. For state-of-the-art comparison, our approach demonstrates superior performance compared to other non-spatial-regularization trackers.*

## 1. Introduction

Visual tracking is one of the core problems in the field of computer vision with a variety of applications. Generic visual tracking is to estimate the trajectory of a target in an image sequence, given only its initial state. It is difficult to design a fast and robust tracker from a very limited set of training samples due to various critical issues in visual tracking, such as occlusion, fast motion and deformation.

Recently, Discriminative Correlation Filter (DCF) [1] has been widely used in visual tracking because of its simplicity and efficiency, and there are many improvements [3, 4, 5, 6, 14, 21] about DCF to address the above mentioned problems. These methods learn a correlation filer from a set of training samples to encode the target's appearance. Nearly all correlation filter based trackers utilize the circulant structure of training samples proposed in work [2]. The circulant structure allows the correlation filter training and target detection computation efficiently. However, this structure also introduces unwanted boundary effects that leads to an inaccurate appearance mode, things get even worse with the growth of searching area.

To address the boundary effects problem, Danelljan et al propose Spatially Regularized Correlation Filters (SRDCF) [5]. The SRDCF introduces penalization to force the correlation filters to concentrate on center of the training patches. This penalization allows the tracker to be trained on a larger area without the effect of background, so the SRDCF can handle some challenging cases such as fast target motion. However, the penalization makes the correlation filters complex to solve. The SRDCF is constructed in the spatial domain, like most other correlation filter based methods, the problem then being transformed into Fourier domain. The result is a complex equation, to solve it, the equation is again transformed into a real-valued one. However, the solution is not the correlation filters we need for detection, so another transformation is needed. These transformations are unacceptable for an online visual tracking situation, especially the second transformation is complex and time consuming. In this work, we revisit the SRDCF, in our formulation, all transformations above are bypassed.

### 1.1. Contributions

In this paper, we propose Faster Spatially Regularized Discriminative Correlation Filters (FSRDCF) for tracking. We construct the FSRDCF from Ridge Regression, the circulant structure of training data matrix in the spatial domain is utilized, besides, we further exploit the circulant structure of regularization matrix in the Fourier domain. With the use of these circulant structures, our method bypasses those transformations in SRDCF [5]. It makes our approach more computation efficient without any signifi-cant degradation in performance and more suitable for online tracking problems.

To validated our approach, we preform comprehensive experiments on three benchmark datasets: OTB-2013 [8] with 50 sequences, OTB-2015 [9] with 100 sequences and VOT2016 [10] with 60 sequences. On OTB-2013 and OTB-2015 datasets, our approach obtains a more than twice faster running speed and a more than third times shorter start-up time than the baseline tracker SRDCF. At the same time, our approach achieves equivalent performance to the SRDCF on all three datasets. For state-of-the-art comparison, our approach demonstrates superior performance compared to other non-spatial-regularization trackers.

## 2. Spatially Regularized DCF

Due to the online nature of the tracking problem, the discriminative correlation filter (DCF) based trackers' simplicity, high efficiency and performance become more and more popular in the tracking community. After Bolme et al. [1] first introduced the MOSSE filter, lots of notable improvements [3, 4, 5, 6, 14] are proposed from different aspects to strengthen the correlation filters based trackers. New features have been widely used, such as HOG [12], Color-Name [13] and deep features [14, 15]; feature integration is also used [20]. To address occlusion, part-based trackers [16] are widely adapted.

All these correlation filter based trackers use Fast Fourier Transform (FFT) to significantly reduce the training and detection computational effort on the base of periodic assumption of the training samples. However, the periodic assumption also produced unwanted boundary effects. Galoogahi et al. [17] investigate the boundary effect issue, their method removes the boundary effects by using a masking matrix to allow the size of training patches larger than correlation filters. They use Alternative Direction Method of Multipliers (ADMM) to solve their problem and have to make transitions between spatial and Fourier domain in every ADMM iteration, which increasing the tracker's computational complexity. To get rid of those transitions, Danelljan et al. [5] propose the spatially Regularized Correlation filters (SRDCF), they introduce a spatial weight function to penalize the magnitude of the correlation filter coefficients, and use Gauss-Seidel method to solve the filters, in this way, both the boundary effects and the transitions in [17] are avoided. However, the SRDCF formulized in the spatial domain is first transformed to Fourier domain to get a complex equation, and then the complex equation is transformed to a real-valued one to use the Gauss-Seidel method to solve it. The Gauss-Seidel method's result have to be transformed again and then get the correlation filters we need. The work [18] also use a spatial regularization like the SRDCF, but they use a different way to solve the problem. When they get the complex equation, they split the complex value into real part and imaginary part, and reconstruct the problem to a real-valued one, then derive a simplified inverse method to get a closed-form solution, however, the simplified inverse operation still have a relatively high computational complexity. In our proposal, we apply a spatial regularization like [5, 18], different form [5, 18],by exploit circulant structure of train samples in the spatial domain and regularization matrix in the Fourier domain, we have no problem transitions from spatial to Fourier or from complex equation to real-valued equation, correlation filters needed are solved directly.

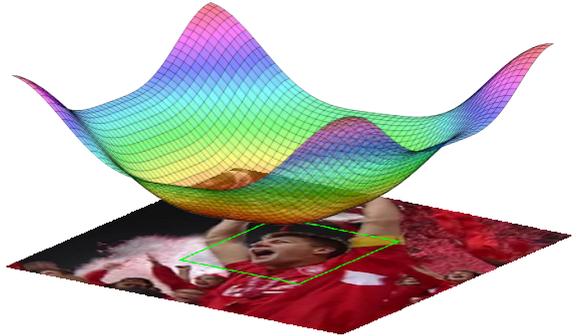

Figure 1: visualization of spatial regularization function $w$ over training patch. $w$ has a strong penalize on marginal area and wake on central area. We also choose $w$ to be even to produce real-valued DFT coefficients.

### 2.1. Standard SRDCF Training and Detection

The way to handle the boundary effects in the SRDCF [5] is most popular in literature [5, 6, 19]. So we give some details of the convolution filters training and Detection after introduced the regularization. In this section, we use the term convolution, because the SRDCF are modeled with convolution instead of correlation, we will give some key differences used in our proposal in section 3 and 4.

Convolution filters $f$ are learned from a set of training samples $\{(x_k, y_k)\}_{k=1}^{t}$. Every training sample $x_k \in \mathbb{R}^{d \times M \times N}$ consists of a d-channel feature map with spatial size of $M \times N$ extracted from a training image patch. We use $x_k^l$ to represent the $l$th feature layer of $x_k$. $y_k$ is the optimal convolution output corresponding to training sample $x_k$. The Spatially Regularized Correlation Filters (SRDCF) is obtained from the convex problem,

$$\min_{f} \sum_{k=1}^{t} \alpha_k \left\| S_f(x_k) - y_k \right\|^2 + \sum_{l=1}^{d} \left\| w \odot f^l \right\|^2. \quad (1)$$

Where the $\alpha_k \geq 0$ is the weight of every training sample $x_k$, spatial regularization is introduced by $w$, which is a Gaussian shaped function with smaller values in center area and bigger values in marginal area, $\odot$ denotes the element-wise multiplication. $S_f(x_k)$ is the convolution function,

$$S_f(x_k) = \sum_{l=1}^{d} x_k^l * f^l. \quad (2)$$

Where $*$ denotes the circular convolution. With the use of Parseval's theorem and convolution property, the Eq. 1 is transformed into Fourier domain,

$$\min_{\hat{f}} \sum_{k=1}^{t} \alpha_k \left\| \sum_{l=1}^{d} \hat{x}_k^l \odot \hat{f}^l - \hat{y}_k \right\|^2 + \sum_{l=1}^{d} \left\| \frac{\hat{w}}{MN} * \hat{f}^l \right\|^2. \quad (3)$$

Where the hat denotes Discrete Fourier Transformed (DFT) of a variable, For convenience, all variables in Eq. 3 are vectorized, convolution is transformed into matrix multiplication,

$$\min_{\hat{\mathbf{f}}} \sum_{k=1}^{t} \alpha_k \left\| \sum_{l=1}^{d} \mathcal{D}(\hat{\mathbf{x}}_\mathbf{k}^\mathbf{l}) \hat{\mathbf{f}}^\mathbf{l} - \hat{\mathbf{y}}_\mathbf{k} \right\|^2 + \sum_{l=1}^{d} \left\| \frac{\mathcal{C}(\hat{\mathbf{w}})}{MN} \hat{\mathbf{f}}^l \right\|^2. \quad (4)$$

Here, bold letters are the corresponding variables' vectorization form, $\mathcal{D}(\mathbf{v})$ is a diagonal matrix with the elements of the vector $\mathbf{v}$ in its diagonal. $\mathcal{C}(\hat{\mathbf{w}})$ is a matrix with its rows consist of all of the shift of the vector $\hat{w}$. Eq. 4 is a complex convex problem, because the DFT of a real-valued function is Hermitian symmetric, so the convex problem (4) can transformed into a real-valued one by a unitary matrix $B \in \mathbb{R}^{MN \times MN}$,

$$\min_{\tilde{\mathbf{f}}} \sum_{k=1}^{t} \alpha_k \left\| \sum_{l=1}^{d} D_k^l \tilde{\mathbf{f}}^l - \tilde{\mathbf{y}}_k^l \right\|^2 + \sum_{l=1}^{d} \left\| C \tilde{\mathbf{f}}^l \right\|^2. \quad (5)$$

Here, $D_k^l = B\mathcal{D}(\hat{\mathbf{x}}_k^l) B^{\mathrm{H}}$, $\tilde{\mathbf{f}}^l = B \hat{\mathbf{f}}^l$, $\tilde{\mathbf{y}}_k = B \hat{\mathbf{y}}_k$ and $C = \frac{1}{MN} B\mathcal{C}(\hat{\mathbf{w}})B^{\mathrm{H}}$, where the $^{\mathrm{H}}$ denotes the conjugate transpose of a matrix. Then concatenate all layers of training data and convolution filters, in other words, $\tilde{\mathbf{f}} = ((\tilde{\mathbf{f}}^1)^T, \cdots, (\tilde{\mathbf{f}}^d)^T)^T$ and $D_k = (D_k^1, \cdots, D_k^d)$, Eq. 5 is simplified as,

$$\min_{\tilde{\mathbf{f}}} \sum_{k=1}^{t} \alpha_k \left\| D_k \tilde{\mathbf{f}} - \tilde{\mathbf{y}}_k \right\|^2 + \left\| W \tilde{\mathbf{f}} \right\|^2. \quad (6)$$

Where $W \in \mathbb{R}^{dMN \times dMN}$ is a block diagonal matrix with its diagonal blocks being equal to $C$. letting the derivative of Eq. 6 with respected to $\tilde{\mathbf{f}}$ be zero,

$$(\sum_{k=1}^{t} \alpha_k D_k^{\mathrm{H}} D_k + W^{\mathrm{H}} W) \tilde{\mathbf{f}} = \sum_{k=1}^{t} \alpha_k D_k^{\mathrm{H}} \tilde{\mathbf{y}}_k. \quad (7)$$

Due to the sparsity of $D_k$ and $W$, problem (7) can be efficiently solved by Gauss-Seidel method with the computational complexity of $\mathcal{O}((d+K^2)dMNN_{GS})$, where the $K$ is the number of non-zero entries in $\hat{w}$, the $N_{GS}$ is the number of Gauss-Seidel iterations. If we want to get the solution of Eq. 4, another transformation is needed,

$$\hat{\mathbf{f}}^l = B^{\mathrm{H}} \tilde{\mathbf{f}}^l. \quad (8)$$

The detection method of the SRDCF is same as standard DCF-based trackers,

$$S_f(z) = \mathcal{F}^{-1} \left\{ \sum_{l=1}^{d} \hat{z}^l \odot \hat{f}^l \right\}. \quad (9)$$

Where $\mathcal{F}^{-1}$ denotes the inverse DFT. In the phase of detection, SRDCF use a scaling pool to handle target scale changes and Fast Sub-grid method to refine the detection results.

Excluding the feature extraction, the total computational complexity of SRDCF tracker is $\mathcal{O}(dSMN\log(MN) + SMNN_{NG} + (d+K^2)dMNN_{GS})$ [5]. Here, $S$ denotes the number of scales in the scaling pool, $N_{NG}$ is the number of Newton iterations in sub-grid detection. It's worth noting that the result takes none of the transformations into consideration, such as Eq. 8, especially from Eq. 4 to Eq. 5, which including high dimensional matrix multiplication. In reality, those transformations are time consuming. In our approach, all of them will be bypassed. We'll directly get the correlation filters in Eq. 8.

## 3. Our Approach

We revisit spatially regularization correlation filters for tracking from Ridge Regression viewpoint. In our proposal, problem is solved more directly by exploiting both circulant structure in training data and regularization function.

### 3.1. Faster SRDCF

Our proposal is to find a function $g(z) = \mathbf{f}^{\mathrm{T}} z$ to minimizes the squared error over all training samples $x_k$ and their regression targets $y_k$,

$$\min_{\mathbf{f}} \sum_{k=1}^{t} \alpha_k \left\| \sum_{l=1}^{d} X_k^l \mathbf{f}^l - \mathbf{y}_k^l \right\|^2 + \sum_{l=1}^{d} \left\| \mathcal{D}(\mathbf{w}) \mathbf{f}^l \right\|^2. (10)$$

Where, for simplicity, we let $\mathbf{y}_k^l = \mathbf{y}_k$. In general Ridge Regression problem, each row of $X_k^l$ is a vectorized training sample, here, rows of $X_k^l$ consist of all circular shift of $\mathbf{x}_k^l$, $X_k^l \mathbf{f}^l$ is the correlation between $\mathbf{x}_k^l$ and $\mathbf{f}^l$, it's worth noting that $X_k^l \mathbf{f}^l \neq vec(x_k^l * f^l)$, where $vec(v) = \mathbf{v}$. Now we can directly take derivative of Eq. 10 with respected to $\mathbf{f}$ and let the derivative be zero, then we get,

$$\sum_{k=1}^{t} A_t \mathbf{f}^l = \sum_{k=1}^{t} \mathbf{b}_t. \quad (11)$$

Where

$$A_t = \sum_{l=1}^{d} \left( X_k^{l\,\mathrm{H}} X_k^l + \mathcal{D}(\mathbf{w})^{\mathrm{H}} \mathcal{D}(\mathbf{w}) \right). \quad (12a)$$

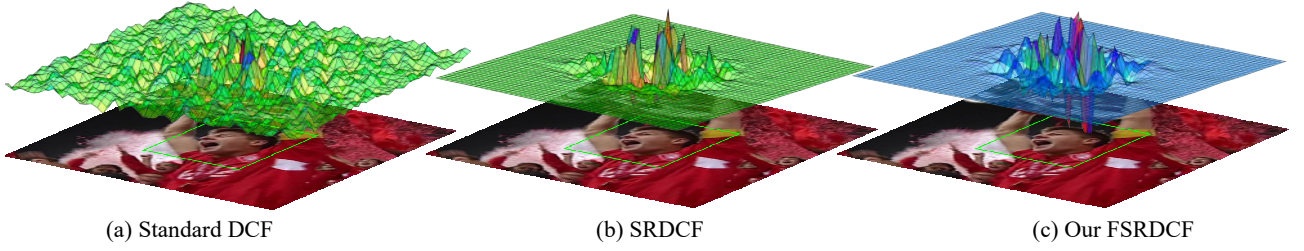

(a) Standard DCF        (b) SRDCF        (c) Our FSRDCF

Figure 2: Visualization of the filter coefficients trained by the standard DCF (a), SRDCF [5] (b) and our FSRDCF (c). The top layer is the learned filter corresponding to the bottom layer training patch. Target is outlined by the green rectangle. Without spatial regularization, high values appear both in target area and background area in filter coefficients trained through standard DCF, this kind of filters are easy to be influenced by mutable background. In SRDCF and our FSRDCF, high values are grouped in the target area, which means filter is concentrate on the target instead of background.

$$\mathbf{b}_t = \sum_{l=1}^{d} X_k^{l\,\mathrm{H}} \mathbf{y}_k^l. \tag{12b}$$

Because all variables in (10) are real-valued, so we use $^\mathrm{T}$ and $^\mathrm{H}$ equivalently. Due to the circulant structure of $X_k^l$, we have [2],

$$X_k^l = \ddot{\mathrm{F}} \mathcal{D}(\hat{\mathbf{x}}_k^l) \ddot{\mathrm{F}}^\mathrm{H}. \tag{13}$$

Where $\ddot{\mathrm{F}} = \mathrm{F} \otimes \mathrm{F}$ is two-dimensional DFT matrix for vectorized two-dimensional signals, F is known as DFT matrix, $\otimes$ denotes the Kronecker product. Both $\ddot{\mathrm{F}}$ and F are constant matrix and unitary. We apply Eq. 13 to Eq. 12,

$$A_t = \sum_{l=1}^{d} \mathrm{F}\bigg(\mathcal{D}(\hat{\mathbf{x}}_k^{l\,*} \odot \hat{\mathbf{x}}_k^l) + \mathrm{F}^\mathrm{H}\mathcal{D}(\mathbf{w})^\mathrm{H}\mathcal{D}(\mathbf{w})\mathrm{F}\bigg)\mathrm{F}^\mathrm{H}. \tag{14}$$

We can see that because of the introduction of regularization $w$, a simple closed-form solution can't be obtained from Eq. 14 like the way in work [2]. So far, we only use the circulant structure of training data matrix $X_k^l$ in the spatial domain. From now on, we will further exploit the circlulant structure of the regularization matrix $\mathcal{D}(\mathbf{w})$.

From Eq. 13, we can know that a spatially circluant matrix can be diagonalized by the matrix of $\ddot{\mathrm{F}}$ in Fourier domain, however, from another point of view, we can also get,

$$\ddot{\mathrm{F}}^\mathrm{H} X_k^l \ddot{\mathrm{F}} = \mathcal{D}(\hat{\mathbf{x}}_k^l). \tag{15}$$

The first row of $X_k^l$ is equal to $\mathcal{F}^{-1}(\hat{\mathbf{x}}_k^l)$, all other rows are the circular shifts of $\mathcal{F}^{-1}(\hat{\mathbf{x}}_k^l)$. So if we have a diagonal matrix, then we can transform it to a circulant matrix by $\ddot{\mathrm{F}}$. In Eq. 14, $\mathcal{D}(\mathbf{w})$ is a real-valued diagonal matrix, so we have

$$\mathrm{F}^\mathrm{H}\mathcal{D}(\mathbf{w})^\mathrm{H}\mathcal{D}(\mathbf{w})\mathrm{F} = \mathrm{F}\mathcal{D}(\mathbf{w})^\mathrm{H}\mathcal{D}(\mathbf{w})\mathrm{F}^\mathrm{H} = \mathrm{R}^\mathrm{H}\mathrm{R}. \tag{16}$$

Where R is circulant matrix constructed from $\mathcal{F}(\mathbf{w})$, where $\mathcal{F}$ denotes DFT. For a real-valued function, unitary DFT and IDFT have the same results, here, we treat $\mathcal{D}(\mathbf{w})$ as a spatial domain signal, so we use DFT instead of IDFT. If we choose a real-valued even regularization function $w$, therefore, $\mathcal{F}^{-1}(\mathbf{w})$ is a real-valued vector, then we will get a real-valued regularization matrix R. Applying Eq. 16 to Eq. 11, we have,

$$\sum_{k=1}^{t}\sum_{l=1}^{d} \mathrm{F}\bigg(\mathcal{D}(\hat{\mathbf{x}}_k^{l\,*} \odot \hat{\mathbf{x}}_k^l) + \mathrm{R}^\mathrm{H}\mathrm{R}\bigg)\ddot{\mathrm{F}}^\mathrm{H}\mathbf{f}^l \\ = \sum_{k=1}^{t}\sum_{l=1}^{d} \mathrm{F}\mathcal{D}(\hat{\mathbf{x}}_k^{l\,*})\ddot{\mathrm{F}}^\mathrm{H}\mathbf{y}_k^l \tag{17}$$

Here, we call R the potential circulant structure of regularization function $w$ in the Fourier domain, By using the unitary property, we can further get,

$$\sum_{k=1}^{t}\sum_{l=1}^{d}\bigg(\mathcal{D}(\hat{\mathbf{x}}_k^{l\,*} \odot \hat{\mathbf{x}}_k^l) + \mathrm{R}^\mathrm{H}\mathrm{R}\bigg)(\ddot{\mathrm{F}}\ddot{\mathrm{F}})^\mathrm{H}\hat{\mathbf{f}}^l \\ = \sum_{k=1}^{t}\sum_{l=1}^{d} \mathcal{D}(\hat{\mathbf{x}}_k^{l\,*})(\ddot{\mathrm{F}}\ddot{\mathrm{F}})^\mathrm{H}\hat{\mathbf{y}}_k^l \tag{18}$$

Where $\ddot{\mathrm{F}}\ddot{\mathrm{F}} = (\mathrm{F} \otimes \mathrm{F})(\mathrm{F} \otimes \mathrm{F}) = (\mathrm{FF}) \otimes (\mathrm{FF})$ is a permutation matrix. To simplify Eq. 18, we define $\hat{\mathbf{f}} = \big((\hat{\mathbf{f}}^l)^\mathrm{T}, \cdots, (\hat{\mathbf{f}}^d)^\mathrm{T}\big)^\mathrm{T}$, $\hat{\mathbf{x}}_k = \big((\hat{\mathbf{x}}_k^l)^\mathrm{T}, \cdots, (\hat{\mathbf{x}}_k^d)^\mathrm{T}\big)^\mathrm{T}$, then problem (18) can be equivalently expressed as,

$$\sum_{k=1}^{t}\bigg(\mathcal{D}(\hat{\mathbf{x}}_k^* \odot \hat{\mathbf{x}}_k) + \mathrm{B}^\mathrm{H}\mathrm{B}\bigg)\mathcal{P}(\hat{\mathbf{f}}) = \sum_{k=1}^{t} \hat{\mathbf{x}}_k^* \odot \mathcal{P}(\hat{\mathbf{y}}_k^l). \tag{19}$$

Where B is a $dMN \times dMN$ block diagonal matrix with each diagonal block being equal to R, $\mathcal{P}(\cdot)$ is a permutation function according to $\ddot{\mathrm{F}}\ddot{\mathrm{F}}$. In section 4, we will find that we just need to find the solution of $\mathcal{P}(\hat{\mathbf{f}}) = \hat{\mathbf{f}}_p$ instead of $\hat{\mathbf{f}}$, so what we really used equation is,

$$\sum_{k=1}^{t}\bigg(\mathcal{D}(\hat{\mathbf{x}}_k^* \odot \hat{\mathbf{x}}_k) + \mathrm{B}^\mathrm{H}\mathrm{B}\bigg)\hat{\mathbf{f}}_p = \sum_{k=1}^{t} \hat{\mathbf{x}}_k^* \odot \mathcal{P}(\hat{\mathbf{y}}). \tag{20}$$

Because we use the same regression targets for all frames, so we define $\hat{\mathbf{y}} = \big((\hat{\mathbf{y}}^1)^\mathrm{T}, \cdots, (\hat{\mathbf{y}}^d)^\mathrm{T}\big)^\mathrm{T}$. Eq. 20 defines a $dMN \times dMN$ linear system of equations, its coefficients matrix is real-valued, so we can solve it directly. We can see that for each frame what we need to do is element-wise multiplication. The regularization part $B^H B$ and regression targets part $\mathcal{P}(\hat{y})$ is constant for all frames.

In Eq. 20, if we choose $B^H B = \lambda I$, which is equivalent to choose a regularization matrix $w$ with all elements being equal to $\sqrt{\lambda}$, we get a standard DCF problem. In this paper, $w$ is a real-valued even Gaussian shaped function, as shown in figure 1, profiting from its smooth property, we get a sparse coefficients matrix for problem (20), which makes significant difference for optimization. The Gauss-Seidel method is used to solve Eq. 20, obviously, the coefficients matrix is symmetric and positive defined, so the converges is guaranteed.

## 4. Our Tracking Framework

In this section, we describe our tracking framework according to the Faster Spatially Regularized Discriminative Correlation Filters proposed in section 3.

### 4.1. Training

At the first frame, to give a better initial point for Gauss-Seidel methods, we obtain a precise solution of Eq. 20 by,

$$\hat{\mathbf{f}}_{p1} = \left(\mathcal{D}(\hat{\mathbf{x}}_1^* \odot \hat{\mathbf{x}}_1) + B^H B\right)^{-1} \left(\hat{\mathbf{x}}_1^* \odot \mathcal{P}(\hat{y})\right). \quad (21)$$

In the subsequent frames, the starting point in current frame (time $t$) is the optimization results in last frame (time $t-1$). For simplicity, we redefine $A_t$ and $\mathbf{b}_t$ in Eq. 11 as,

$$A_t = \sum_{k=1}^{t} \left(\mathcal{D}(\hat{\mathbf{x}}_k^* \odot \hat{\mathbf{x}}_k) + B^H B\right). \quad (22a)$$

$$\mathbf{b}_t = \sum_{k=1}^{t} \hat{\mathbf{x}}_k^* \odot \mathcal{P}(\hat{y}). \quad (22b)$$

Eq. 20 can be rewrite as $A_t \hat{\mathbf{f}}_p = \mathbf{b}_t$, we split $A_t$ into data part $D_t$ and regularization part $B^H B$, split $\mathbf{b}_t$ into $\mathbf{d}_t$ and $\mathcal{P}(\hat{y})$, then we update our model by,

$$A_t = (1-\gamma)D_{t-1} + \gamma\mathcal{D}(\hat{\mathbf{x}}_t^* \odot \hat{\mathbf{x}}_t) + B^H B. \quad (24a)$$

$$\mathbf{b}_t = \left((1-\gamma)\mathbf{d}_{t-1} + \gamma\hat{\mathbf{x}}_t^*\right)\odot\mathcal{P}(\hat{y}). \quad (24b)$$

Where $D_1 = \mathcal{D}(\hat{\mathbf{x}}_1^* \odot \hat{\mathbf{x}}_1), \mathbf{d}_1 = \hat{\mathbf{x}}_1^*$. In Eq. 24, $B^H B$ and $\mathcal{P}(\hat{y})$ are constant during model updating. We just need to precompute once for a sequence. To get new correlation filters $\hat{\mathbf{f}}_p$, a fixed $N_{GS}$ numbers of Gauss-Seidel iterations are conducted after model updating Eq. 24.

### 4.2. Detection

At the detection stage, according Eq. 10, the location of the target is estimated by finding the peak correlation between correlation filters $\hat{\mathbf{f}}_p$ and new feature maps,

$$\max \sum_{l=1}^{d} X_t^l \mathbf{f}^l = \max \sum_{l=1}^{d} x_t^l \star f^l. \quad (25)$$

Where $\star$ denotes the correlation operator. For the sake of computational efficiency, we use convolution instead of correlation,

$$\max \sum_{l=1}^{d} x_t^l \star f^l = \max \sum_{l=1}^{d} x_t^l \ast (-f^l). \quad (26)$$

Where $-$ denotes $180°$ rotation operator. both $x_t$ and $f$ are real-valued, so their DFT is Hermitian symmetric. Computing Eq. 26 in Fourier domain, finally we get,

$$\max \mathcal{F}^{-1}\left(\sum_{l=1}^{l} \hat{\mathbf{x}}_t^l \odot \hat{\mathbf{f}}_p^l\right). \quad (27)$$

So in Eq. 19, we directly solve $\mathcal{P}(\hat{\mathbf{f}}) = \hat{\mathbf{f}}_p$ instead of $\hat{\mathbf{f}}$. At detection stage, we also use the scaling pool technique in paper [20] and Fast Sub-grid Detection in work [5].

## 5. Experiments

To validate our proposal, we make comprehensive experiments, and report results on three benchmark datasets: OTB2013, OTB2015 and VOT2016.

### 5.1. Details and Parameters

To make a fair comparison with SRDCF [5], we use most of the parameters used in SRDCF throughout all our experiments. Because we need get real-valued DFT coefficients of the regularization function $w$ and use Matlab as our implementation tool, so we reconstructed $w$ from function $w(m,n) = \mu + \eta(\mathbf{m}/\sigma_1)^2 + \eta(\mathbf{n}/\sigma_2)^2$, where $[\sigma_1, \sigma_2] = \beta[P, Q]$, $P \times Q$, is the size of target. $\mathbf{m} = [-(M/2) : (M-2)/2)]$, $\mathbf{n} = [-(N/2) : (N-2)/2)]$. In our experiments, $\beta$ is set to 0.8. From the construction process of $w$, we know that the minima is not at the center of $w$. To remove this bias, we set all feature map size to be odd, and do same shifts during image patch extracting for training and detection.

### 5.2. Baseline Comparison

We do comprehensive comparison between our approach and the baseline tracker SRDCF [5]. Accuracy, robustness and speed are taken into consideration. In this section, all experiments are performed on a standard desktop computer with Intel Core i5-6400 processor. For the baseline tracker, we use the Matlab implementation provided by authors.

|  | Attribute-based evaluation |  |  |  |  |  |  |  |  | overall |  |
|---|---|---|---|---|---|---|---|---|---|---|---|
|  | SV | OCC | DEF | OV | IPR | OPR | BC | LR | IV | AUC | OP |
| SRDCF | 59.5 | 62.7 | 63.5 | 55.5 | 57.3 | 60.4 | 58.7 | 42.6 | 57.6 | 63.0 | 78.9 |
| FSRDCF | 61.8 | 63.6 | 65.4 | 54.9 | 59.6 | 63.0 | 61.9 | 42.1 | 58.6 | 64.4 | 81.6 |
| SRDCF | 56.9 | 55.7 | 54.7 | 46.1 | 54.6 | 55.1 | 58.4 | 48.1 | 60.9 | 59.9 | 73.1 |
| FSRDCF | 56.4 | 56.1 | 54.3 | 45.6 | 56.4 | 56.3 | 59.0 | 46.8 | 60.8 | 59.5 | 73.4 |

Table 1. Comparison with baseline tracker on OTB-2013 (the first two rows of data) and OTB-2015. The results in the table are based on success plot and all reported in percent. Mean OP is a threshold-based evaluation. Attribute-based evaluations are performed on 11 attribute sub-datasets, results are reported for 9 attributes in the consideration of space.

|  | OTB-2013 | | OTB-2015 | |
|---|---|---|---|---|
|  | SRE | TRE | SRE | TRE |
| SRDCF | 0.569 | 0.647 | 0.542 | 0.613 |
| FSRDCF | 0.557 | 0.650 | 0.522 | 0.607 |

Table 2. Robustness evaluation comparison on OTB-2013 and OTB-2015 datasets. Both trackers achieve equivalent results in TRE on OTB-2015 dataset. Due to the using of bigger derivation for regularization function ($\beta = 0.8$), our tracker is more sensitive to background when preforming spatial robustness evaluation.

### 5.2.1 Accuracy and robustness comparison

Under benchmark datasets OTB-2013 and OTB-2015, we follow the protocol proposed in [8]. One-Pass Evaluation (OPE), Temporal Robustness Evaluation (TRE) and Spatial Robustness Evaluation (SRE) are performed. TRE runs trackers on 20 different length sub-sequences segmented from the original sequences, SRE run trackers with 12 different initializations constructed from shifted or scaled ground truth bounding box. After running the trackers, we report the overall results using the area under the curve (AUC) based on success plot and mean overlap precision (OP). Besides, attribute-based evaluation results are also reported. The OP is calculated as the percentage of frames where the intersection-over-union overlap with the ground truth exceeds a threshold of 0.5. Attributes are including scale variation (SV), occlusion (OCC), deformation (DEF), fast motion (FM), in-plane-rotation (IPR), out-plane-rotation (OPR), background cluster (BC) and low resolution (LR), illumination variation (IV), out of view (OV), motion blur (MB).

Table 1 shows OPE results on OTB-2013 and OTB-2015. On OTB-2013 dataset, for clarity, we reported 9 attribute-based evaluation results. For overall performance, our approach outperforms the baseline tracker by 1.1%, 2.7% in AUC and OP respectively on OTB-2013 dataset and achieves equivalent performance both in AUC and OP score on OTB-2015 dataset. For attribute-based evaluation, our method wins in most attribute sub-datasets on OTB-2013 dataset, on OTB-2015 dataset, the performance of both trackers have no significant difference on all sub-datasets and finally keep the overall performance equivalently. Table 2 shows the robustness evaluation results. Except for our tracker is a little bit more sensitive to the background due to bigger derivation parameters, we think two trackers have the same robustness performance.

|  | EAO | A | R |
|---|---|---|---|
| SRDCF | 0.1848 | 0.51 | 2.27 |
| FSRDCF | 0.1834 | 0.51 | 2.13 |

Table 3. Overall comparison with baseline tracker on VOT-2016 dataset. Both tracker achieve equivalent performance in accuracy and robustness.

|  | CM | IC | OCC | SC | MC |
|---|---|---|---|---|---|
| SRDCF | 0.1886 | 0.1593 | 0.1561 | 0.1546 | 0.1232 |
| FSRDCF | 0.1822 | 0.1425 | 0.1669 | 0.1411 | 0.1219 |

Table 4. Attribute based comparison with baseline tracker on VOT-2016 dataset. The EAO is reported on 5 attribute sub-datasets. Both trackers obtain very close EAO scores.

|  | OTB-2013 | | OTB-2015 | |
|---|---|---|---|---|
|  | Speed | Start-up | Speed | Start-up |
| SRDCF | 5.7 | 1.27 | 5.3 | 1.36 |
| FSRDCF | 11.4 | 0.51 | 11.1 | 0.50 |

Table 5. The comparison of speed and start-up time on OTB-2013 and OTB-2015. The trackers' speed is reported in fps, start-up time is reported in second (s). Our approach run more than twice fast as the baseline tracker SRDCF, and more than three times shorter in start-up time on OTB-2015 dataset.

Under the benchmark dataset VOT-2016, we follow the work [10] and evaluate our tracker on 60 videos. We report accuracy (A), robustness (R), expected average overlap (EAO). These measures evaluate a tracker from different aspects. The accuracy is the average overlap between the predicted and ground truth bounding boxes during successful tracking periods. Robustness measures how many times the tracker loses the target during tracking. EAO is proposed in work [21] to overcome the drawback that AO [8] is affected by the sequence lengths. Attribute-based evaluations are also performed. Videos in VOT-2016 dataset are labeled with 5 attributes: camera motion (CM), illumination change (IC), occlusion (OCC), size change (SC), motion change (MC). Based on the labeled sub-datasets, we perform an attribute analysis.

Table 3 shows the A, R and EAO scores over all the 60 videos on VOT-2016 datasets. The differences in EAO and A score between both trackers are less than 0.01, at the same time, we can also find that both trackers get a very close EAO score on 5 attribute sub-datasets in Table 4. From the aspect of robustness, our approach outperforms the baseline tracker by 0.14.

### 5.2.2 Speed comparison

To compare the speed of our approach to the baseline tracker SRDCF, we evaluate both trackers on OTB-2013 and OTB-2015 datasets. To get a more intuitive and meaningful results, we report our experiments results in frames per second (fps) instead of equivalent filters operations (EFO) [11], as a result, we run both trackers on the same computer. In Table 5, we give the overall speed performance results on OTB-2013 and OTB-2015 datasets. To have a more detailed comparison, we list the speed of both trackers on every video in OTB-2013 dataset in Table 6. All speed evaluation above is excluding the start-up time of trackers, however, the start-up time may be very important, especially when trackers need to be reinitialized frequently. We define the interval from the beginning of an algorithm to the first model being learned from the first frame as the tracker's start-up time. We list both trackers start-up time performance on OTB-2013 dataset in Table 6. Because both trackers use almost the same model parameters, the speed of trackers has close relationship with the size of targets, so in Table 6 we also give the size of the feature map both trackers used, the size is computed as $\sqrt{M \times N}$.

| Sequences | CarDark | | Car4 | | David | | David2 | | Sylvester | | Trellis | | Fish | | Mhyang | | Soccer | | Matrix | | Ironman | | Deer | |
|---|---|---|---|---|---|---|---|---|---|---|---|---|---|---|---|---|---|---|---|---|---|---|---|---|
| Size | 25.8 | | 50 | | 50 | | 30 | | 50 | | 50 | | 50 | | 50 | | 50 | | 40.0 | | 50 | | 50 | |
| SRDCF | 9 | 0.8 | 4 | 1.6 | 4 | 1.6 | 11 | 1.4 | 4 | 1.6 | 4 | 1.4 | 4 | 1.4 | 4 | 1.6 | 4 | 1.6 | 6 | 1.0 | 4 | 1.6 | 4 | 1.4 |
| FSRDCF | 16 | 0.3 | 7 | 0.8 | 8 | 0.8 | 22 | 0.2 | 8 | 0.8 | 10 | 0.5 | 10 | 0.5 | 8 | 0.8 | 7 | 0.8 | 11 | 0.5 | 7 | 0.8 | 9 | 0.5 |
| Ratio | 1.8 | 0.4 | 1.8 | 0.5 | 2.0 | 0.5 | 2.0 | 0.1 | 2.0 | 0.5 | 2.5 | 0.4 | 2.5 | 0.4 | 2.0 | 0.5 | 1.8 | 0.5 | 1.8 | 0.5 | 1.8 | 0.5 | 2.3 | 0.4 |
| Skating1 | | Shaking | | Singer1 | | Singer2 | | Coke | | Bolt | | Boy | | Dudek | | Crossing | | Couple | | Football | | Jogging1 | | Jogging2 | |
| 50 | | 50 | | 50 | | 50 | | 50 | | 40 | | 38 | | 50 | | 29 | | 39 | | 33 | | 50 | | 50 | |
| 4 | 1.5 | 4 | 1.6 | 4 | 1.3 | 4 | 1.5 | 4 | 1.4 | 6 | 1.0 | 7 | 0.9 | 4 | 1.6 | 12 | 0.6 | 7 | 0.9 | 10 | 0.6 | 4 | 1.3 | 4 | 1.5 |
| 8 | 0.6 | 7 | 0.8 | 11 | 0.2 | 8 | 0.6 | 10 | 0.5 | 14 | 0.4 | 11 | 0.4 | 7 | 0.8 | 21 | 0.2 | 12 | 0.3 | 18 | 0.2 | 12 | 0.2 | 8 | 0.6 |
| 2.0 | 0.4 | 1.8 | 0.5 | 2.8 | 0.2 | 2.0 | 0.4 | 2.5 | 0.4 | 2.3 | 0.4 | 1.6 | 0.4 | 1.8 | 0.5 | 1.8 | 0.3 | 1.7 | 0.3 | 1.8 | 0.3 | 3.0 | 0.2 | 2.0 | 0.4 |
| Doll | | Girl | | Walking2 | | Walking | | Fleetface | | Freeman1 | | Freeman3 | | Freeman4 | | David3 | | Jumping | | CarScale | | Skiing | | Dog1 | |
| 48 | | 37 | | 50 | | 44 | | 50 | | 25 | | 12 | | 15 | | 50 | | 33 | | 33 | | 27 | | 43 | |
| 4 | 1.4 | 8 | 0.8 | 4 | 1.3 | 5 | 1.2 | 4 | 1.6 | 9 | 0.7 | 17 | 0.4 | 11 | 0.6 | 4 | 1.3 | 10 | 0.7 | 9 | 0.6 | 7 | 0.9 | 6 | 1.0 |
| 8 | 0.5 | 14 | 0.2 | 13 | 0.2 | 10 | 0.4 | 8 | 0.8 | 18 | 0.3 | 32 | 0.2 | 19 | 0.3 | 12 | 0.2 | 18 | 0.3 | 19 | 0.2 | 11 | 0.4 | 9 | 0.6 |
| 2.0 | 0.4 | 1.8 | 0.3 | 3.3 | 0.2 | 2.0 | 0.3 | 2.0 | 0.5 | 2.0 | 0.4 | 1.9 | 0.5 | 1.7 | 0.5 | 3.0 | 0.2 | 1.8 | 0.4 | 2.1 | 0.5 | 1.6 | 0.5 | 1.5 | 0.6 |
| Suv | | MotorRolling | | MountainBike | | Lemming | | Liquor | | Woman | | Faceocc1 | | Faceocc2 | | Basketball | | Football | | Subway | | Tiger1 | | Tiger2 | |
| 50 | | 50 | | 50 | | 50 | | 50 | | 46 | | 50 | | 50 | | 50 | | 44 | | 31 | | 50 | | 50 | |
| 4 | 1.5 | 4 | 1.7 | 4 | 1.9 | 4 | 1.6 | 4 | 1.7 | 5 | 1.1 | 4 | 1.5 | 4 | 1.7 | 4 | 1.7 | 5 | 1.3 | 10 | 0.7 | 4 | 1.8 | 4 | 1.8 |
| 9 | 0.6 | 7 | 0.8 | 7 | 0.8 | 8 | 0.6 | 8 | 0.6 | 14 | 0.2 | 7 | 0.8 | 8 | 0.8 | 8 | 0.6 | 9 | 0.6 | 19 | 0.2 | 7 | 0.8 | 7 | 0.8 |
| 2.3 | 0.4 | 1.8 | 0.5 | 1.8 | 0.4 | 2.0 | 0.4 | 2.0 | 0.4 | 2.8 | 0.2 | 1.8 | 0.5 | 2.0 | 0.5 | 2.0 | 0.4 | 1.8 | 0.5 | 1.9 | 0.3 | 1.8 | 0.4 | 1.8 | 0.4 |

Table 6. The details of speed and start-up time of our approach FSRDCF and the baseline tracker SRDCF on OTB-2013 dataset. Each row including 5 sub-rows, corresponding to sequences name, size of single feature layer, speed (fps) and start-up time (second) of SRDCF, speed and start-up time of FSRDCF, the ratio of FSRDCF to SRDCF in speed and start-up time respectively. Each sequence including 2 sub-column, the first is speed, the second is start-up time.

In Table 5, our tracker runs at 11.4fps, 11.1fps on OTB-2013 and OTB-2015 datasets respectively, the baseline tracker SRDCF obtains 5.7fps, 5.3fps correspondingly. Our tracker runs at a twice fast speed as the SRDCF on both datasets. From the aspect of the start-up time, our tracker obtains a mean value of 0.50s on OTB-2015 comparing to SRDCF's 1.36s. With faster running speed and shorter start-up time, our tracker is more suitable for online tracking applications than the SRDCF. In Table 6, we list the trackers' speed and start-up time against to the feature size both trackers used on each video in OTB-2013 dataset. We found that our tracker outperform much on those videos which have a larger feature size, in our experiments, feature size is restricted to a maxim size of $50 \times 50$. However, when trackers handle big targets or utilize a high dimensional features, our tracker can give a relatively much higher running speed.

### 5.3. OTB-2013 and OTB-2015 datasets

Finally, we perform a comprehensive comparison with 18 recent state-of-art trackers: DLSSVM [22], SCT4 [23], MEEM [25], KCF [2], DSST [25], SAMF [20], LCT [7], MIL [26], IVT [27], TLD [28], ASLA [29], L1APG [30], CSK [31], SCM [32], LOT [33], Frag [34], Struck [35] and the baseline tracker SRDCF [5].

#### 5.3.1 State-of-the-art comparison

We show the results of comparison with state-of-the-art trackers on OTB-2013 and OTB-2015 datasets over 100 videos in Table 7, only the results for the top 7 trackers are reported in consideration of space. The results are presented in mean overlap precision (OP) and ranking according to performance on OTB-2015 dataset. In Table 7, we also give the speed of top 7 trackers in fps. The best results on both datasets are obtained by our tracker with mean OP of 81.6%. and 73.4%, outperforming the best non-spatial regularization trackers by 8.4% and 6.3% respectively. From the perspective of running speed, our approach runs at 11.1 frames per second, which is more than twice faster than the tracker ranking the second. Our tracker gets a better balance between accuracy and efficiency.

Figure 4 shows the success plots on OTB-2013 and OTB-2015 datasets. The success plot shows the ratios of successful frames at the intersection-over-union based threshold varied from 0 to 1. The trackers are ranked acc-ording to the area under the curve (AUC) and displayed in the legend. Our tracker ranks the first on OTB-2013 with a AUC of 64.4%, outperforming the best non-spatial regularization tracker by 5.1%, and ranks the second with a AUC of 59.5%, outperforming the best non-spatial-regularization tracker by 4.3% on OTB-2015.

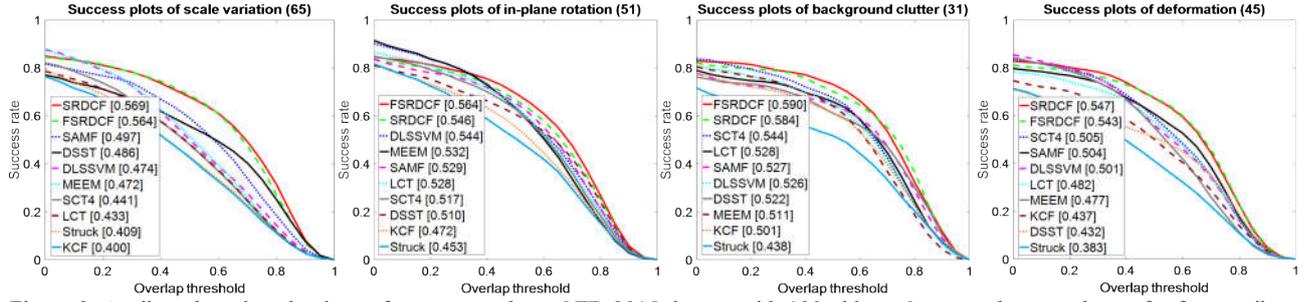

Figure 3. Attribute-based evaluations of our approach on OTB-2015 dataset with 100 videos. Success plots are shown for four attribute sub-datasets, number in bracket of each plot title is the videos in corresponding sub-dataset. Only top 10 trackers are displayed in each plot. Our tracker demonstrates superior performance compared to other non-spatial regularization trackers.

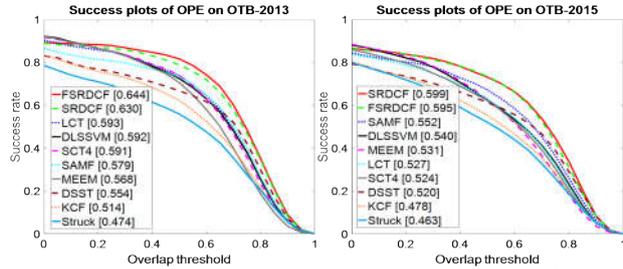

Figure 4. Success plots showing a comparison with state-of-the-art trackers on OTB-2013 (left) and OTB-2015 (right) datasets. For clarity, only the top 10 methods are displayed. Our FSRDCF ranks the first on OTB-2013 and the second on OTB-2015.

#### 5.3.2 Robustness comparison

Like in section 5.2.1, we perform SRE and TRE to compare the robustness of our tracker to the state-of-the-art trackers. Figure 5 shows success plots for SRE and TRE on OTB-2015 dataset with 100 videos. Our approach outperforms the best non-spatial-regularization tracker 1% and 2% in SRE and TRE respectively.

#### 5.3.3 Attribute based comparison

We perform attribute-based evaluations of our approach on OTB-2015 and compare to other state-of-the-art trackers. Our approach wins on 10 attribute sub-datasets compared to other non-spatial-regularization trackers, Figure 3 shows the success plots of 4 different attributes on OTB-2015 dataset. Due to the using of spatial regularization, the spatially regularized trackers can learn more discriminative filters and detect targets from a lager area than standard DCF, so our tracker have big advantages in situations such as occlusion, background cluster and fast motion over other trackers without spatial regularization.

|  | DSST | SCT4 | DLSSVM | MEEM | LCT | SAMF | SRDCF | FSRDCF |
|---|---|---|---|---|---|---|---|---|
| OTB-2013 | 66.7 | 73.9 | 72.5 | 70.6 | 73.8 | 73.2 | 78.9 | 81.6 |
| OTB-2015 | 61.3 | 62.0 | 62.4 | 62.7 | 62.9 | 67.1 | 73.1 | 73.4 |
| Speed | 29.4 | 32.2 | 9.5 | 8.2 | 20.3 | 16.8 | 5.3 | 11.1 |

Table 7. State-of–the-art trackers comparison on OTB-2013 and OTB-2015 datasets using mean overlap precision (in percent). Speed is reported in fps according performance on OTB-2015 dataset. The best two results are shown in red and blue respectively. Our approach achieves the best results on both benchmark datasets and have a balanced perform on accuracy and speed.

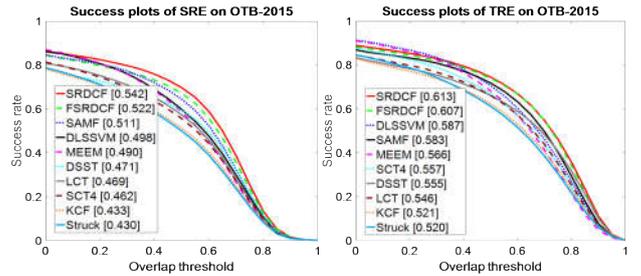

Figure 5. Robustness to initialization comparison on the OTB-2015 dataset. Success plots for both SRE and TRE are shown, our tracker achieves state-of-the-art performance.

## 6. Conclusion

We propose Faster Spatially Regularized Discriminative Correlation Filters (FSRDCF) to efficiently learn a spatially regularized correlation filer which addresses the limitation of standard DCF. The use of circulant structure of data matrix in the spatial domain and circulant structure of regularization function in the Fourier domain significantly simplify the problem construction and solving. In our approach, both problem construction and solving are in the spatial domain. We validated our approach on three benchmark datasets. On OTB-2013 and OTB-2015 datasets, our approach obtains a more than twice faster running speed and a more than third times shorter start-up time than the baseline tracker SRDCF. At the same time, our approach achieves equivalent performance to the SRDCF on all three datasets. For state-of-the-art comparison, our approach demonstrates superior performance compared to other non-spatial-regularization trackers.

# References

[1] D. S. Bolme, J. R. Beveridge, B. A. Draper, and Y. M. Lui. Visual object tracking using adaptive correlation filters. In *CVPR*, 2010.

[2] J. F. Henriques, R. Caseiro, P. Martins, and J. Batista. High speed tracking with kernelized correlation filters. *PAMI*, 2015.

[3] T. Liu, G. Wang, and Q. Yang. Real-time part-based visual tracking via adaptive correlation filters. In *CVPR*, 2015.

[4] M. Danelljan, G. Hager, F. S. Khan, and M. Felsberg. Accurate scale estimation for robust visual tracking. In *BMVC*, 2014.

[5] M. Danelljan, G. Häger, F. S. Khan, and M. Felsberg. Learning Spatially Regularized Correlation Filters for Visual Tracking. In *ICCV*, 2015.

[6] M. Danelljan, G. Häger, F. S. Khan, and M. Felsberg. Adaptive Decontamination of the Training Set: A Unified Formulation for Discriminative Visual Tracking. In *CVPR*, 2016.

[7] C. Ma, X. Yang, C. Zhang, and M. Yang. Long-term correlation tracking. In *CVPR*, 2015.

[8] Y. Wu, J. Lim, and M.-H. Yang. Online object tracking: A benchmark. In *CVPR*, 2013.

[9] Y. Wu, J. Lim, and M.-H. Yang. Object tracking benchmark. *PAMI*, 2015.

[10] M. Kristan, Aleš Leonardis, Jiři Matas, Michael Felsberg, and et al. The visual object tracking vot2016 challenge results. In *ECCV Workshop*, 2016.

[11] M. Kristan, R. Pflugfelder, A. Leonardis, J. Matas, and et al. The visual object tracking vot2014 challenge results. In *ECCV Workshop*, 2014.

[12] N. Dalal and B. Triggs. Histograms of oriented gradients for human detection. In *CVPR*, 2005.

[13] J. van de Weijer, C. Schmid, J. J. Verbeek, and D. Larlus. Learning color names for real-world applications. In *TIP*, 2009.

[14] C. Ma, J.-B. Huang, X. Yang, and M.-H. Yang. Hierarchical convolutional features for visual tracking. In *ICCV*, 2015.

[15] H. Nam and B. Han. Learning multi-domain convolutional neural networks for visual tracking. In *CVPR*, 2016.

[16] T. Liu, G. Wang, and Q. Yang. Real-time part-based visual tracking via adaptive correlation filters. In *CVPR*, 2015.

[17] H. K. Galoogahi, T. Sim, and S. Lucey. Correlation filters with limited boundaries. In *CVPR*, 2015.

[18] Z. Cui, S. Xiao, J. Feng, S. Yan. Recurrently Target-Attending Tacking. In *CVPR*, 2016.

[19] M. Danelljan, A. Robinson, F. S. Khan, M. Felsberg. Beyond Correlation Filter Operations for Visual Tracking. In *ECCV*, 2016.

[20] Y. Li and J. Zhu. A scale adaptive kernel correlation filter tracker with feature integration. In *ECCV Workshop*, 2014.

[21] S. Gladh, M. Danelljan, F. S. Khan, M. Felsberg. Deep motion features for visual tracking. In *ICPR*, 2016.

[22] J. Ning, J. Yang, S. Jiang, L. Zhang and et al. Object Tracking via Dual Linear Structure SVM and Explicit Feature Map. In *CVPR*, 2016.

[23] J. Chi, H. Jin Chang, J. Jeong, Y. Demiris and et al. Visual Tracking Using Attention-Modulated Disintegration and integration. In *CVPR*, 2016.

[24] J. Zhang, S. Ma, and S. Sclaroff. MEEM: robust tracking via multiple experts using entropy minimization. In *ECCV*, 2014.

[25] M. Danelljan, G. Häger, F. S. Khan, and M. Felsberg. Accurate scale estimation for robust visual tracking. In *BMVC*, 2014.

[26] B. Babenko, M.-H. Yang, and S. Belongie. Visual tracking with online multiple instance learning. In *CVPR*, 2009.

[27] D. Ross, J. Lim, R.-S. Lin, and M.-H. Yang. Incremental learning for robust visual tracking. *IJCV*, 77(1):125–141, 2008.

[28] Z. Kalal, J. Matas, and K. Mikolajczyk. P-n learning: Bootstrapping binary classifiers by structural constraints. In *CVPR*, 2010.

[29] X. Jia, H. Lu, and M.-H. Yang. Visual tracking via adaptive structural local sparse appearance model. In *CVPR*, 2012.

[30] C. Bao, Y. Wu, H. Ling, and H. Ji. Real time robust l1 tracker using accelerated proximal gradient approach. In *CVPR*, 2012.

[31] J. F. Henriques, R. Caseiro, P. Martins, and J. Batista. Exploiting the circulant structure of tracking-by-detection with kernels. In *ECCV*, 2012.

[32] W. Zhong, H. Lu, and M.-H. Yang. Robust object tracking via sparsity-based collaborative model. In *CVPR*, 2012.

[33] S. Oron, A. Bar-Hillel, D. Levi, and S. Avidan. Locally orderless tracking. In *CVPR*, 2012.

[34] A. Adam, E. Rivlin, and Shimshoni. Robust fragments-based tracking using the integral histogram. In *CVPR*, 2006.

[35] S. Hare, A. Saffari, and P. Torr. Struck: Structured output tracking with kernels. In *ICCV*, 2011.